\documentclass[11pt]{article}

\usepackage[margin=1in]{geometry}
\usepackage{setspace}
\usepackage{cite}
\usepackage{amsmath,amssymb,amsfonts}
\usepackage{graphicx, float}
\usepackage{textcomp}
\usepackage{xcolor}
\usepackage{csquotes}
\usepackage{subcaption}
\usepackage{multicol}
\usepackage{lipsum}
\usepackage[bookmarks=false]{hyperref}
\usepackage{wrapfig}
\def\BibTeX{{\rm B\kern-.05em{\sc i\kern-.025em b}\kern-.08em
    T\kern-.1667em\lower.7ex\hbox{E}\kern-.125emX}}
    
\usepackage{tabularx}
\usepackage{soul}
\usepackage{algorithm}
\usepackage{algpseudocode}
\algblock{Input}{EndInput}
\algnotext{EndInput}
\algblock{Output}{EndOutput}
\algnotext{EndOutput}
\usepackage{tikz}
\usetikzlibrary{matrix}

\begin{document}
\title{The Application of Transformer-Based Models for Predicting Consequences of Cyber Attacks}

\author{
    Bipin Chhetri and Akbar Siami Namin \\
    Department of Computer Science \\
    Texas Tech University \\
    Lubbock, TX, USA \\
    \texttt{bipin.chhetri@ttu.edu, akbar.namin@ttu.edu}
}

\date{}

\maketitle

\begin{center}
\textit{This work has been published in COMPSAC Symposium on Emerging Advances in Technologies \& Applications (EATA 2025), IEEE COMPSAC 2025 IEEE International Conference on Computers, Software, \& Applications, Toronto, Canada, July 8-11, 2025}
\end{center}

\begin{abstract}

Cyberattacks are increasing, and securing against such threats is costing industries billions of dollars annually. Threat Modeling, that is, comprehending the consequences of these attacks, can provide critical support to cybersecurity professionals, enabling them to take timely action and allocate resources that could be used elsewhere. Cybersecurity is heavily dependent on threat modeling, as it assists security experts in assessing and mitigating risks related to identifying vulnerabilities and threats. Recently, there has been a pressing need for automated methods to assess attack descriptions and forecast the future consequences of the increasing complexity of cyberattacks. This study examines how Natural Language Processing (NLP) and deep learning can be applied to analyze the potential impact of cyberattacks by leveraging textual descriptions from the MITRE Common Weakness Enumeration (CWE) database. We emphasize classifying attack consequences into five principal categories: Availability, Access Control, Confidentiality, Integrity, and Other. This paper investigates the use of Bidirectional Encoder Representations from Transformers (BERT) in combination with Hierarchical Attention Networks (HANs) for Multi-label classification, evaluating their performance in comparison with conventional CNN and LSTM-based models. 
Experimental findings show that BERT achieves an overall accuracy of $0.972$, far higher than conventional deep learning models in multi-label classification.  HAN outperforms baseline forms of CNN and LSTM-based models on specific cybersecurity labels. However, BERT consistently achieves better precision and recall, making it more suitable for predicting the consequences of a cyberattack.
 
\end{abstract}


\section{Introduction}

Cyberattacks are becoming increasingly frequent and sophisticated, affecting critical infrastructure, cloud services, and healthcare systems on an unprecedented scale. In August 2023, Amazon Web Services (AWS) experienced a significant Distributed Denial of Service (DDoS) attack, targeting Amazon S3 and resulting in more than 155 million requests per second. The incident caused an eight-hour outage, which affected access to vital services and highlighted the vulnerability of the cloud infrastructure to major cyber threats \cite{aws_ddos}. In another incident on that date, Google Cloud services experienced an alarming DDoS attack driven by HTTP/2, which peaked at 398 million requests per second. Unlike all previous events \cite{google_ddos}, this incident was the biggest Layer 7 assault known to date.

Beyond cloud services, the healthcare industry has also experienced numerous instances of cyberattacks. In February 2024, Change Healthcare  \cite{change_healthcare} experienced a significant ransomware attack, compromising the medical and personal records of about 190 million people.  The attack revealed private information, including names, phone numbers, social security numbers, and medical histories, resulting in one of the most significant health breaches, which caused significant operational challenges \cite{change_healthcare}.

Threat modeling is an indispensable process within the field of cybersecurity, providing a structured methodology for security experts to assess and understand the potential threats and vulnerabilities inherent to complex systems \cite{swiderski2004threat}. The application of machine learning in cybersecurity has attracted a myriad of interests and is widely utilized to address security-related problems such as intrusion detection \cite{kocher2021machine, bakro2023improved,muneer2024critical}, malware analysis \cite{mitra2023survey, akhtar2022malware}, anomaly detection \cite{ajala2024leveraging,haji2021attack} and vulnerability detection \cite{bahaa2024db},\cite{gopali2022vulnerability}.
Researchers continually devise new methods and strategies to enhance machine learning's capabilities in tackling cybersecurity problems. However, the rapid advancement of technology has provided criminals with the opportunity to target frequent and sophisticated threats that were previously unattainable. The attacks on AWS \cite{aws_ddos} and Google \cite{google_ddos} serve as a lesson of how modern cyber threats can be carried out with unprecedented speed and efficiency. 

Cyberattacks are increasing in frequency and complexity. Therefore, protecting systems is now vital. 
By allowing for a more efficient and context-aware examination of whole text sequences \cite{vaswani2017attention}, transformer models have greatly improved Natural Language Processing (NLP). Traditional models, such as Recurrent Neural Networks (RNNs) and Convolutional Neural Networks (CNNs), consider an input in a specific order or focus on local features. On the other hand, transformers utilize self-attention methods to determine the importance of each word in a given phrase, and therefore have evolved into a fundamental component of modern NLP systems.

Inspired by a previous study by Datta et al. \cite{datta2022can}, where the application of CNN and LSTM models for multi-label classifications was explored to predict consequences of cyberattacks, this paper introduces a hybrid approach that combines these deep architectures with attention mechanisms and ensemble learning through transformer-based models such as BERT transformer models and Hierarchical Attention Networks (HAN) with the objective of enhancing the accuracy of predicting consequences of cyberattacks. The goal is to improve the accuracy of consequence prediction, particularly for consequences modeled through mainstream threat modeling schema such as CIR (i.e., Confidentiality, Integrity, and Availability). This paper specifically focuses on predicting multiple labels, including Availability, Access Control, Confidentiality, Integrity, and Other associated with each textual description drawn from MITRE's Common Weakness Enumeration (CWE) dataset. \cite{MITRE_CWE} The real-world results of our study demonstrate that both the Hierarchical Attention Networks (HANs) deep neural network model and the BERT transformer model outperform the state-of-the-art CNN and RNN-based models. 


This study addresses the critical challenge of predicting the consequences of cyberattacks that have implications for end-users, a task complicated by the multi-label nature of cybersecurity vulnerabilities and the limitations of existing models in handling such complexity. Traditional methods, such as CNN, RNN, and knowledge graph-based approaches, often overlook the complex relationships in textual data, despite the models' need to understand the broader context. 
By examining these issues, the study helps to enhance threat modeling techniques and improve the accuracy of predictions about potential outcomes following a cyberattack incident. The paper makes the following key contributions:
\begin{itemize}
    \item[--] \textbf{Transformer-based BERT multi-label Classification.} We introduce a transformer-based BERT model fine-tuned for predicting five key labels representing consequences of cyberattacks, such as Availability, Access Control, Confidentiality, Integrity, and Other, using textual descriptions from the MITRE CWE dataset \cite{MITRE_CWE}.
    \item[--] \textbf{Hierarchical Attention Networks (HAN).} The study demonstrates the effectiveness of HANs in capturing document-level semantics, enhancing the precision of multi-label classification for two labels, i.e., Access control and Integrity.
    \item[--] \textbf{Improved Performance.} Empirical results show that the BERT-based model significantly outperforms traditional CNN-LSTM architectures, achieving superior accuracy and F1-scores across all labels. Notably, BERT achieves the best performance overall, while HAN excels in specific categories (i.e., two labels).
    \item[--] \textbf{Practical and Scalable Solution.} This work offers a scalable alternative to knowledge graph-based methods, providing a streamlined approach suitable for real-world applications in cybersecurity.
\end{itemize}

The paper is organized as follows. Section \ref{sec:relatedwork} introduces the related work. Section \ref{sec:background} provides a brief overview of the technical methodologies adopted in this paper. Section \ref{sec:models} presents the architecture and implementation of the proposed models. Section \ref{sec:experimental_setup} provides the details of the dataset preprocessing techniques and experimental setup. The methodology employed in this paper is described in Section \ref{sec:methodology}. Section \ref{sec:analysis_results} discusses the analysis and results. Section \ref{sec:Comparison} compares the proposed approach with previous work. 
Section \ref{sec:conclusion} concludes the paper and outlines future research directions.

\section{Related work}
\label{sec:relatedwork}

 Various deep learning models, such as Long Short-Term Memory 
 networks (LSTMs), Recurrent neural network (RNN), Convolutional 
 Neural Networks (CNNs), and hybrid models have been extensively 
 applied in cybersecurity for tasks like anomaly  detection \cite{wei2023lstm}, Entity recognition \cite{gasmi2024lstm} and threat mitigation \cite{yadav2024mitigation}. These models often face challenges when analyzing complex textual data inherent in cybersecurity contexts \cite{gasmi2019information}. With its self-attention methods, transformer-based models are known as a major development as they provide better performance on several NLP tasks, including text categorization. This section provides an overview of current deep learning techniques, Hybrid-LSTM and RNN, and transformer-based models that are initially used for various applications mitigating cyberattacks.
 
Hui et al. \cite{hui2020extraction} proposed a fusion model that combines the attention mechanism with Bi-LSTM and BERT, which significantly outperforms single-model approaches. Their model achieved an accuracy of 89.52\% on short medical text from Traditional Chinese Medicine (TCM) medical records, highlighting the potential of combining contextual embeddings with sequential learning. Their work is limited to short medical texts, which may not generalize well to longer, more complex documents. Additionally, their study does not address challenges in multi-label texts, which may not be generalized constraints such as cybersecurity.

Mahdaouy et al. \cite{ra/maliciousurl/BERT} used a pre-trained BERT-based encoder designed for detecting and classifying suspicious or malicious domain names and URLs. For pre-training, the Masked Language Modeling (MLM) objective was applied to a large multilingual dataset that included URLs, domain names, and Domain Generation Algorithm (DGA) data. The authors evaluated the performance of their models on various classification tasks, including phishing, malware detection, and DNS tunneling. The result showed DomURLs\_BERT outperforms state-of-the-art character-based deep learning models across multiple datasets. The study does not thoroughly address its vulnerability to adversarial attacks. 

Cheng et al.\cite{ra/cyberbully/HAN} proposed a Hierarchical Attention Network in cyberbullying detection (HANCD) that captures temporal behavior patterns of cyberbullying detection. A real-world Instagram dataset from \cite{ra/2015detection} demonstrates that incorporating temporal dynamics improves performance by 5.3\% compared to the Hierarchical Attention Network with Temporal Features (HANT). The study is limited to Instagram, and the approach may not be readily generalized to other social media platforms, such as Facebook and X.

Xu et al. \cite{ra/networkintrusion} introduced a dual-domain intrusion detection (DDT) model that combines Temporal Convolutional Networks (TCN) to extract local and global features, addressing the growing complexity of network attacks. Their work not only focuses on intrusion detection but also highlights the potential of Transformer-based models in cybersecurity applications. The DTT model exhibits improvements in F1-score ranging from 0.6 to 6.8 on the NCCI dataset and from 0.4 to 3.5 on the NUB dataset compared to other models. The model requires extensive pre-training, making it less suitable for real-time intrusion detection scenarios.

Nguyen et al. \cite{ra/flowbased/BERT} proposed an approach to improve the domain adaptation capability of Network Intrusion Detection Systems (NIDS) using Natural Language Processing (NLP) and the BERT framework. The network traffic flows were organized as sequences, similar to sentences in language. The authors trained the BERT model using the Masked Language Modeling (MLM) task and then fine-tuned it with a linear layer and softmax output for intrusion detection. It achieved an F1-score of $0.877$ and an accuracy of $0.916$. The model achieved positive results across different domains. The model still relies on labeled data, which limits its scalability and applicability.

Deep learning models have been extensively utilized for multi-label text classification tasks. Early works, such as Kim \cite{kim2014cnn}, employed a simple Convolutional Neural Network (CNN) with pre-trained word2vec embeddings to predict sentiment and classify questions. This study highlighted the effectiveness of CNNs in text classification and laid the groundwork for subsequent hybrid models. Later methods combined CNNs and Recurrent Neural Networks (RNNs) to leverage the strengths of both types of architecture. 

In cybersecurity, knowledge graphs have emerged as a promising tool for modeling complex relationships among vulnerabilities, threats, and consequences.
Han et al. \cite{han2023knowledgegraph} constructed a knowledge graph using CWE data to predict relationships between entities and classify cybersecurity consequences. These efforts have automated processes such as assigning CWE-IDs to CWE entries and have shown potential in organizing cybersecurity information more effectively. Their reliance on structured graph representations contrasts with this study, which leverages deep learning techniques to predict multi-label outcomes without the need for knowledge graphs.


Recent research has also explored machine learning techniques for predicting attack consequences in cybersecurity. Datta et al. \cite{datta2023mlattacks} applied machine learning models to analyze 93 attack descriptions and predict 50 potential consequences from an end-user perspective. Similarly, Dass et al. \cite{dass2023hmm} utilized Hidden Markov Models (HMMs) to anticipate attack consequences from the victim’s perspective, providing a proof-of-concept for spoofing attacks. These studies underscore the importance of accurately modeling and predicting the impacts of attacks, yet they rely on traditional techniques that may not scale effectively to larger datasets.

The work presented in this paper fills this gap by utilizing transformer-based models for multi-label classification, which allows us to predict cybersecurity attack outcomes exclusively from textual descriptions. Previous research has primarily employed knowledge graphs or basic machine learning techniques. This study, on the other hand, utilizes advanced attention and transformer-based architectures, such as BERT and HAN, to enhance the accuracy and scalability of multi-label classification in cybersecurity settings.

\section{Preliminary}
\label{sec:background}



\subsection{Transformers}

The Transformer-based models, introduced by Vaswani et al. \cite{vaswani2017attention}, have dramatically changed deep learning, as a self-attention mechanism is able to capture long-range dependencies without dealing with the inherent problems (e.g., vanishing gradient) that often occur in recurrent or convolutional neural networks. Unlike RNNs and LSTMs, which process input sequentially, transformers are able to operate in parallel, resulting in huge efficiency and scalability benefits. The multi-head self-attention is used to give the model the chance to assess the relevance of different words in a sentence by itself. The model architecture utilizes "position encoding" to preserve word order information. This architecture has led to the most recent breakthroughs in multiple areas, such as natural language processing\cite{bajaj2024gtmicro,rahali2023end}, time series forecasting\cite{woo2024unified}, and computer vision\cite{dubey2024transformer,chen2024survey}, with models including BERT, GPT, and T5 demonstrating deeper contextual knowledge. Despite its impressive performance, the Transformer's high computational cost and memory usage severely limit its capacity. This is the driving ongoing research into the domain of more efficient attention techniques, such as sparse and linear attention mechanisms. As optimizations continue to improve in terms of scalability, transformers are at the heart of modern AI research and applications.


\subsection{Hierarchical Attention Network (HAN)}

Hierarchical Attention Networks (HANs) is an effective model used in document-level representations, particularly for long-text classification tasks \cite{yang2016hierarchical}. Unlike conventional deep learning models, HAN leverages attention mechanisms at multiple levels words and sentences to capture contextual meaning while focusing on the most relevant parts of the text. Given the multi-label nature of cybersecurity vulnerability classification, HAN is a suitable choice because it can highlights informative contents from textual descriptions. 

The architecture comprises word-level and sentence-level attention mechanisms that aggregate meaningful representations for classification. The model begins by processing tokenized words through an embedding layer, such as GloVe, followed by a bidirectional Gated Recurrent Unit (GRU) that generates context-aware word embeddings. A word-level attention mechanism assigns weights to words based on their importance in the classification task. The sentence-level attention follows a similar approach, where word representations are aggregated into sentence embeddings, processed again by a bidirectional GRU, and refined through attention to highlight significant sentences. The final classification layer maps the attended sentence representations to multiple labels using a fully connected dense layer with a softmax or sigmoid activation function. This hierarchical approach enables HAN to efficiently capture contextual relationships within text and improve classification accuracy.

\section{Transformer-based Model Architectures for Predicting Consequences of Cyberattacks}
\label{sec:models}

\subsection{The BERT Model}

The foundation of this model is based on the pre-trained ``\textit{bert-base-uncased}''  \cite{huggingface_bert_base_uncased}, which is derived from the BERT architecture initially proposed in \cite{devlin2019bert}. This version is one of the first pre-trained models that was made public and fine-tuned for classification tasks where the output can be any consequence (e.g., integrity, confidentiality, etc.). Because it can utilize contextual embeddings, it is a suitable choice for many natural language processing applications. A pre-trained \textbf{BERT} (Bidirectional Encoder Representations from Transformers) model (\texttt{bert-base-uncased}) was employed, comprising the following components:

\begin{enumerate}
    
    \item \textbf{BERT Encoder}. The encoder comprises 12 transformer layers that generate contextually rich embeddings for input tokens, leveraging self-attention mechanisms to learn relationships between words within the text.
    
    \item \textbf{Input Layer}. The model processes tokenized sequences derived from cybersecurity vulnerability descriptions, with a maximum sequence length set of 256 tokens.
    
    \item \textbf{Dropout Layer}. Applied with a probability of 0.3 to prevent overfitting.
    
    \item \textbf{Linear Layer}. Maps the 768-dimensional pooled output of the BERT model to five output neurons corresponding to the multi-label targets. A \textbf{sigmoid} activation function was applied to each of the five output neurons to predict each label's probability independently. The sigmoid activation function is commonly used in binary classification tasks that require a value between 0 and 1, making it ideal for predicting the likelihood of an event occurring\cite{rumelhart1986learning}.
\end{enumerate}

\begin{figure*}[!htbp]
    \centering
    \begin{subfigure}[b]{0.45\textwidth}
        \centering
        \includegraphics[width=\linewidth, height=6cm]{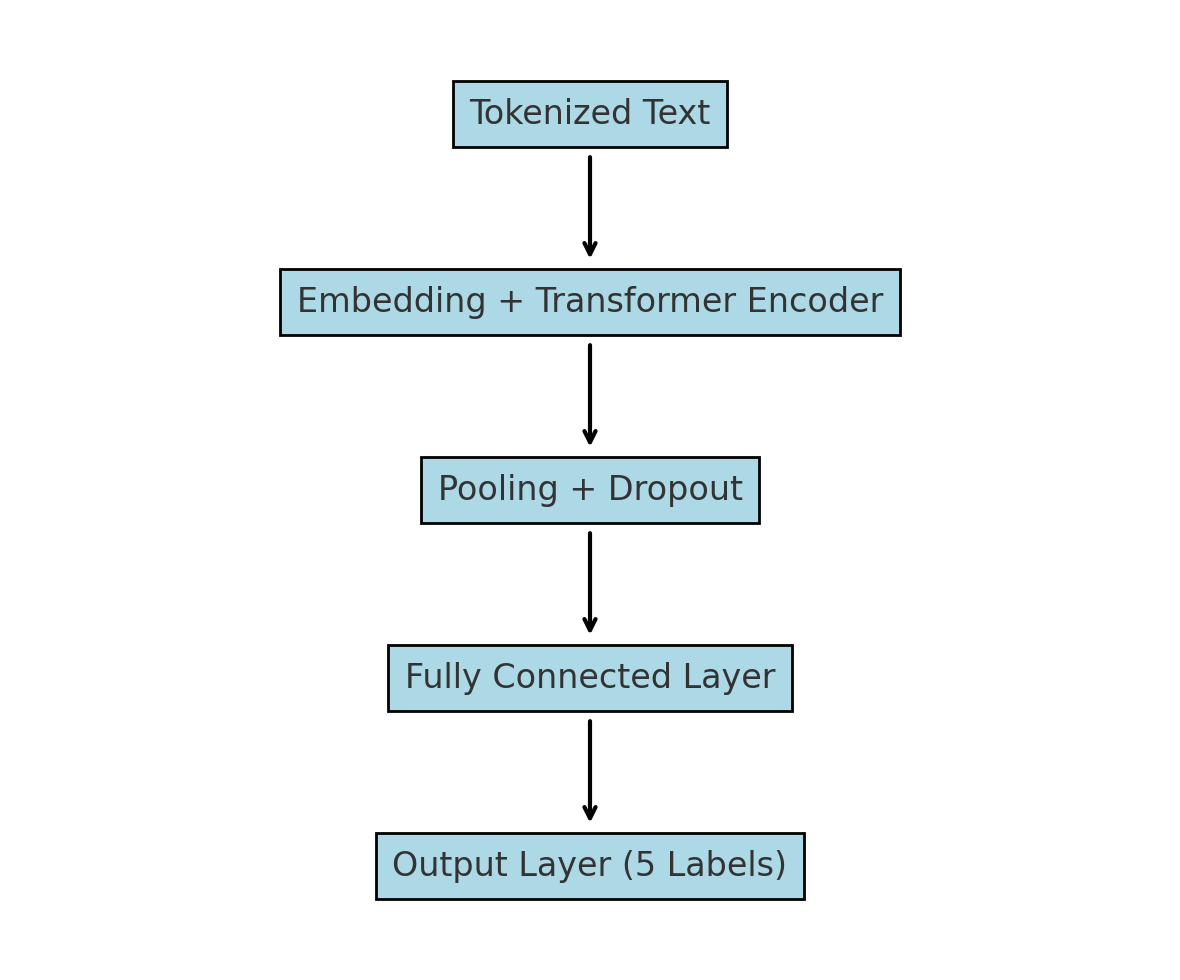}
        \caption{BERT-Based Model Architecture}
        \label{fig:model_architecture}
    \end{subfigure}
    \hspace{0.05\textwidth} 
    \begin{subfigure}[b]{0.45\textwidth}
        \centering
        \includegraphics[width=\linewidth, height=6cm]{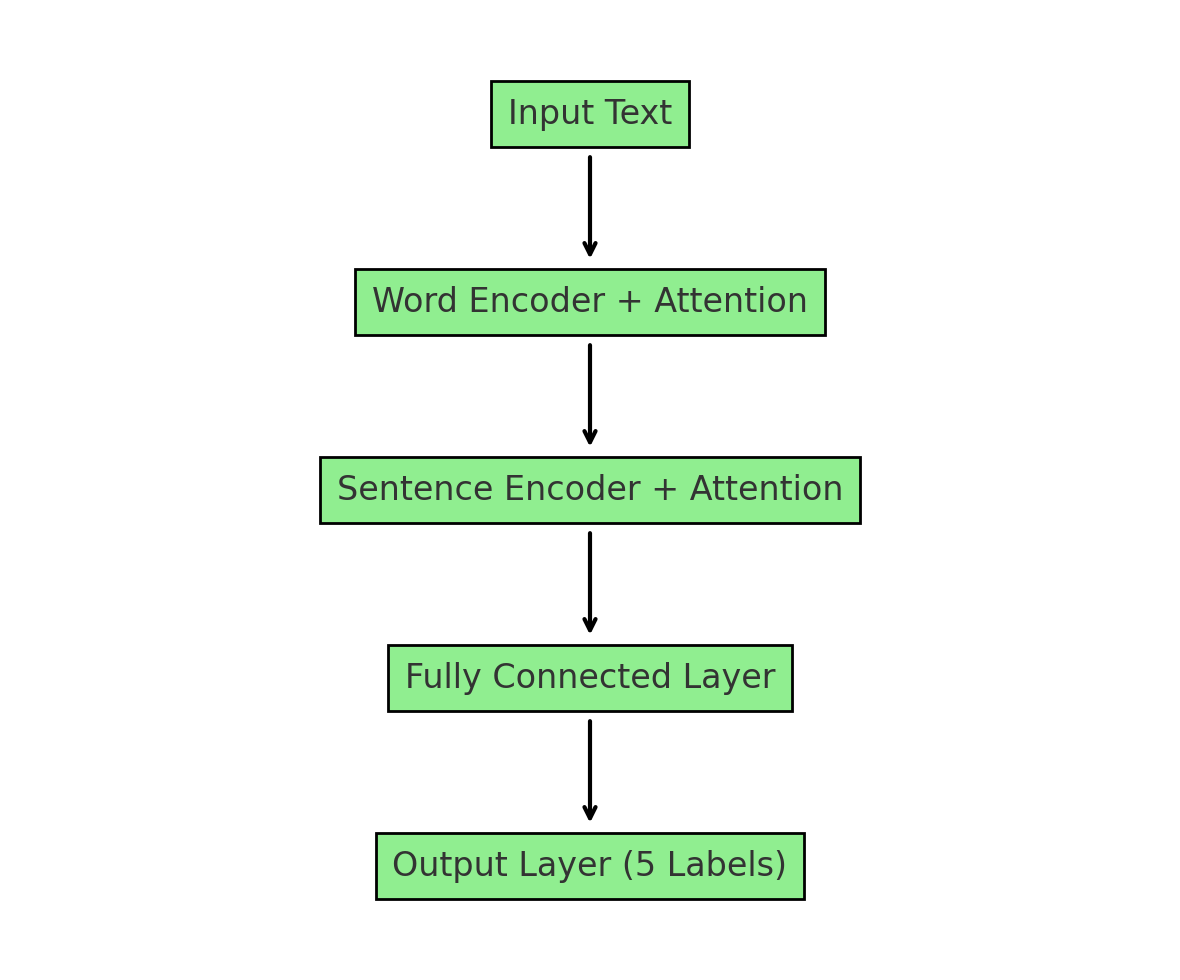}
        \caption{HAN-Based Model Architecture}
        \label{fig:han_model_architecture}
    \end{subfigure}
    \caption{Architectural Diagrams of BERT and HAN Models Used for Predicting Cyberattack Consequences}
    \label{fig:model_architectures}
    \vspace*{-0.15in} 
\end{figure*}

Figure \ref{fig:model_architecture} illustrates the architecture of the BERT transformer model with multi-labeled classification capability to predict the consequences of cyber attacks. 

\subsection{Hyper-parameter Tuning for BERT}
Hyper-parameter tuning was crucial in fine-tuning the model for optimal performance. The following key hyper-parameter tunings were applied to the BERT model:

\begin{itemize}
    \item[--] \textbf{Maximum Sequence Length}: 256 tokens, ensuring that longer descriptions are fully captured without truncation.
    \item[--] \textbf{Batch Size}: A batch size of 32 was utilized during both training and validation.
    \item[--] \textbf{Learning Rate}: A learning rate of \(1 \times 10^{-5}\), optimized using the Adam optimizer, was found to be the most effective after testing different values.
\end{itemize}


\subsection{Algorithm for Predicting Consequences with BERT Layer}
Algorithm \ref{alg:bert_multi-label} outlines the BERT-based training and prediction steps. Initially, the dataset is split into training, validation, and testing subsets to ensure fair evaluation. Mini-batches generated from the training data are used to fine-tune the model after it has been initialized with pre-trained weights. Each mini-batch undergoes tokenization using the BERT tokenizer, after which the tokenized input is fed through the encoder layers. A dropout regularization technique is then applied, followed by a fully connected output layer for multi-label classification. During training, the binary cross-entropy loss is calculated and backpropagated to update the model weights. In the prediction phase, the trained model feeds test data through the same pipeline to get probability distributions for all labels.


\begin{algorithm}[t]
\caption{BERT-Based Classification Training Process.}
\label{alg:bert_multi-label}
\begin{algorithmic}[1]
\State \textbf{Input:} Training Dataset 
\State \textbf{Output:} Trained BERT model
\State Split Data: Training, Validation, Testing. 
\Procedure{Train Model}{}
    \State Initialize BERT model with pre-trained weights
    \For{each epoch $e = 1$ to $E$}
        \For{each mini-batch $B \in Training$}
            \State Tokenize $B$
            \State Apply dropout and fully connected layers 

        \EndFor
        \State Evaluate validation performance 
    \EndFor
    \State \textbf{Return} trained BERT model
\EndProcedure

\Procedure{Predict}{Testing Data $X_{test}$}
    \State Tokenize $X_{test}$
    \State Pass tokens through trained BERT model
    \State Apply sigmoid activation 
    \State Convert scores to binary labels using a threshold 
    \State \textbf{Return} predicted labels $\hat{Y}$
\EndProcedure

\end{algorithmic}
\end{algorithm}


\subsection{The HAN Model}

The foundation of this model is a Hierarchical Attention Network (HAN), which is designed to capture document-level structures by leveraging both word-level and sentence-level attention mechanisms. Inspired by the work performed by Yang et al. \cite{yang2016hierarchical}, the HAN model is particularly effective for text classification tasks, as it learns to focus on the most informative words and sentences when making predictions. For this specific task, HAN was adapted for multi-label classification, predicting consequences such as Integrity, Confidentiality, Availability, Access Control, and Other. A Hierarchical Attention Network consists of the following key components:

\begin{enumerate}
    \item {\bf Word Encoder}. A bidirectional Gated Recurrent Unit (Bi-GRU) was employed to learn the contextual representation of each word within a sentence. The embedding layer converts tokenized words into dense vectors, which are then passed through the Bi-GRU.
    
    \item {\bf Word-Level Attention}. An attention mechanism is applied at the word level to assign weights to words based on their importance within the sentence. This allows the model to focus on key terms that contribute most to classification.
    
    \item {\bf Sentence Encoder}. The weighted word representations are aggregated into a sentence vector, which is then passed through another Bi-GRU layer to learn sentence dependencies.
    
    \item {\bf Sentence-Level Attention}. Similar to word-level attention, a sentence-level attention mechanism is employed to assess the significance of each sentence within the overall document, thereby enhancing classification accuracy.
    
    \item {\bf Fully Connected Layer}. The final document representation is passed through a fully connected layer with five output neurons, each representing a multi-label category.
    
    \item {\bf Sigmoid Activation}. A sigmoid activation function is applied to each of the five output neurons to produce independent probabilities for each consequence label.
\end{enumerate}


Figure \ref{fig:han_model_architecture} illustrates the Hierarchical Attention Network architecture for predicting cyber attack consequences.

\subsection{Hyper-parameter Tuning for HAN}
Hyperparameter tuning was crucial for fine-tuning the HAN model to achieve optimal performance. The following key hyperparameters were used:

\begin{itemize}
     \item[--] \textbf{Maximum Sequence Length:} 256 tokens, ensuring adequate representation of cybersecurity vulnerability descriptions.
     \item[--] \textbf{Batch Size:} A batch size of 32 was used for both training and validation.
     \item[--] \textbf{Learning Rate:} \(1 \times 10^{-4}\), optimized using the Adam optimizer.
    \item[--] \textbf{GRU Units:} The GRU layers at the word level and sentence level consist of 64 hidden units.
\end{itemize}

\subsection{Algorithm for Predicting Consequences with HAN Model}

\begin{algorithm}[t]
\caption{HAN-Based multi-label Classification Training.}
\label{alg:han_multi-label}
\begin{algorithmic}[1]
\State \textbf{Input:} Training Dataset 
\State \textbf{Output:} Predicted labels for test data
\State Split Data: Training, Validation, Testing. 
\Procedure{Train Model}{}
    \State Initialize word and sentence Bi-GRU layers with attention mechanisms
    \For{each epoch $e = 1$ to $E$}
        \For{each mini-batch $B \in Training$}
            \State Tokenize and embed words
            \State Encode words using Bi-GRU
            \State Apply attention layer
            \State Pass fully connected layer
        \EndFor
        \State Validate Model performance
    \EndFor
\EndProcedure
\end{algorithmic}
\end{algorithm}


\begin{figure*}[htbp]
    \centering
    \includegraphics[width=\textwidth, height=0.9\textheight, keepaspectratio]{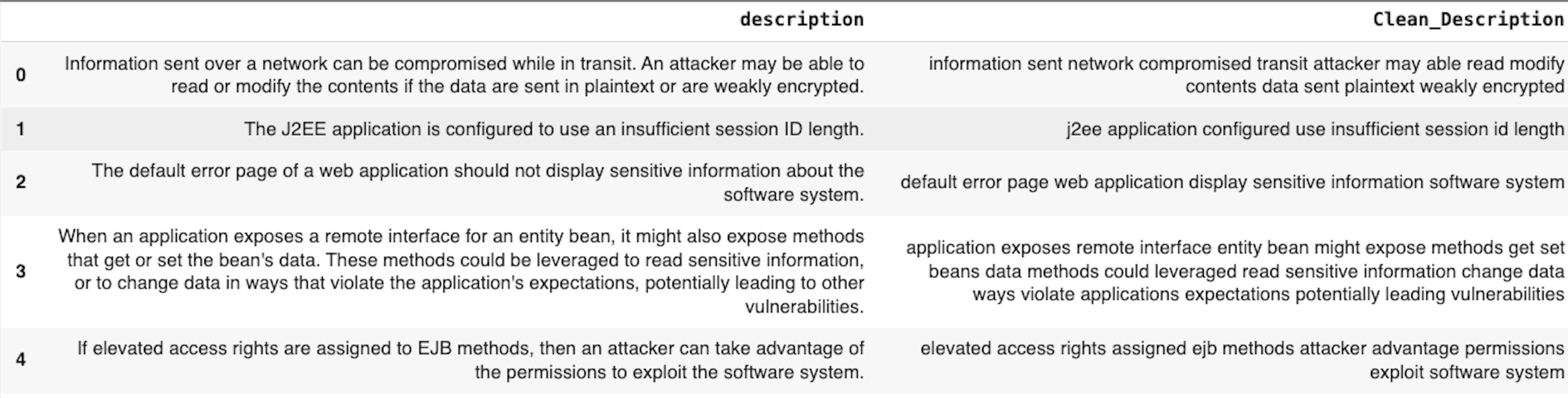}
    \caption{A sample of CWE dataset with the original description and description after cleaning.}
    \label{figure:dataset1A}
        \vspace*{-0.20in}
\end{figure*}

Algorithm \ref{alg:han_multi-label} outlines the HAN-based training and 
prediction steps. The training corpus is first divided into stratified subsets 
to preserve the distribution of multi-label classes. The model is then 
initialized with Bi-GRU encoders at both the word and sentence 
levels, each followed by dedicated attention modules. Pre-trained word vectors 
are used to tokenize and embed input text during each training cycle. The word 
sequences in each sentence are passed through a word-level Bi-GRU, where an 
attention mechanism assigns weights to their contextual embeddings, 
prioritizing semantically relevant tokens. The aggregated weighted word 
embeddings were then used to construct sentence representations, which a sentence-level Bi-GRU processed.

\section{Experimental Setup}
\label{sec:experimental_setup}

\subsection{Data Pre processing}
The dataset used in this study is derived from an enhanced version of the MITRE Common Weakness Enumeration (CWE) dataset \cite{MITRE_CWE}. The CWE database is continuously updated and maintained to reflect newly identified attacks. At the time of data collection, it contained $1,016$ distinct CWE entries. The task is set up as a multi-label classification problem because we need to predict the outcome based on text descriptions, and each CWE entry can be linked to more than one outcome. Initially, there were six labels. However, to balance the dataset, we retained five columns: availability, access control, confidentiality, integrity, and non-repudiation, among others. To ensure a balanced distribution of the multi-label classes across the subsets, the dataset was divided using stratified sampling. After filtering the dataset, \textbf{895 rows} remained, each containing descriptions of cybersecurity vulnerabilities labeled with one or more consequences. The five target labels for this study were: \textbf{Availability}, \textbf{Access Control}, \textbf{Confidentiality}, \textbf{Integrity}, and \textbf{Other}.

\subsection{Data Cleaning}
\label{sec:after preprocessing}

\begin{figure*}[htbp]
    \centering
    \includegraphics[width=\textwidth, height=0.9\textheight, keepaspectratio]{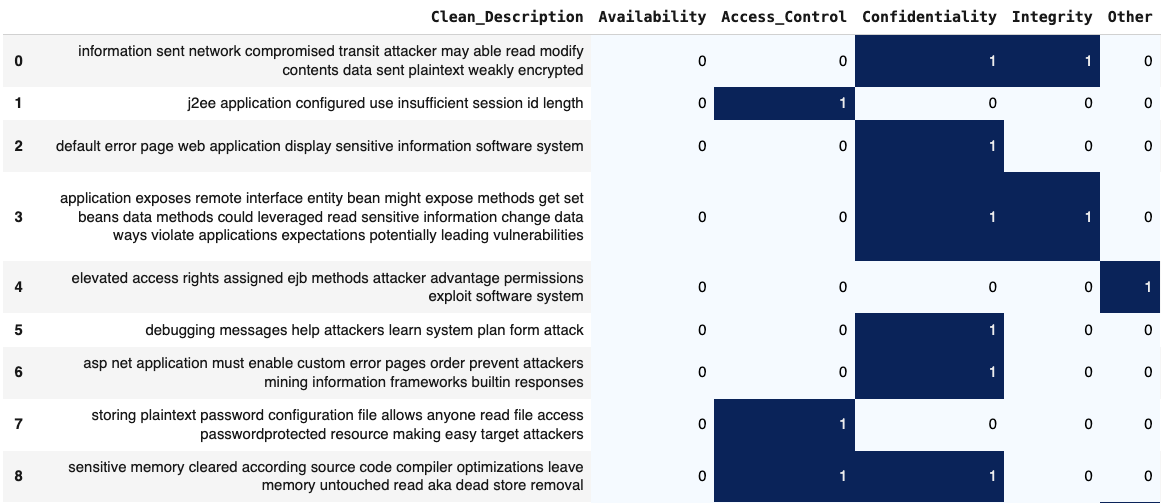}
    \caption{A sample of the CWE dataset with the clean description  and five labels.}
    \label{figure:dataset1B}
        \vspace*{-0.20in}
\end{figure*}

 The initial dataset includes raw text data, which may contain unnecessary symbols, stop words, and inconsistent capitalization. These artifacts needed to be cleaned for efficient processing. Figure \ref{figure:dataset1A} depicts a sample of the CWE dataset before preprocessing of the description and the cleaning stage. Once the necessary preprocessing techniques, such as removing stopwords, tokenization, and normalizing the text, was applied, the dataset was cleaned and made ready for model training.
Prior to training, the following preprocessing steps were applied to the dataset:
\begin{itemize}
    \item[--] \textbf{Text Cleaning}. Descriptions were cleaned using Python-based NLTK (Natural Language Toolkit), removing irrelevant information such as punctuation, stopwords, and non-text symbols. The text was also converted to lowercase to standardize the input.
     \item[--] \textbf{Target Labels}. Only five key target labels were retained (\textit{Availability}, \textit{Access Control}, \textit{Confidentiality}, \textit{Integrity}, \textit{Other}), and redundant fields like \textit{id}, \textit{name}, and \textit{extended\_description} were excluded to focus the dataset on the most relevant information.
\end{itemize}
After the cleaning process and the removal of labels associated with smaller subsets, Figure \ref{figure:dataset1B} depicts a cleaned CWE description paired with its corresponding five multi-label classifications.

\subsection{Tokenization}
The $BertTokenizer$ from the Hugging Face transformer library \cite{huggingface_bert_base_uncased} was used to tokenize the text descriptions. The text was encoded into input tokens with padding and truncation applied to ensure a uniform sequence length $MAX\_LEN = 256$.

\subsection{Performance Metrics}
Metrics such as \textbf{accuracy}, \textbf{precision}, \textbf{recall}, and \textbf{F1-score} were calculated for both micro and macro averages. These metrics were computed using the Scikit-learn library \cite{scikit_learn}. Validation was carried out at the end of each epoch to monitor performance. The model performance evaluation details are as follows: 
\begin{itemize}
    \item \textbf{Accuracy}: The proportion of correctly classified labels over the total number of labels.
    \item \textbf{Precision}: The ratio of true positives to all predicted positives, relevant in cases of imbalanced data.
    \item \textbf{Recall}: The ratio of true positives to all actual positives, indicating the model’s ability to capture relevant instances.
    \item \textbf{F1-Score}: The harmonic mean of precision and recall, calculated for both micro and macro averages.
\end{itemize}

\section{Methodology}
\label{sec:methodology}

\subsection{Training Process}
The training loop involved minimizing the binary cross-entropy loss between predicted and actual labels. Model checkpoints were saved at each epoch if the validation loss improved, and early stopping was employed to prevent overfitting. 
A validation set was used at the end of each epoch to monitor performance and prevent overfitting. The model was trained on an \textbf{NVIDIA A100 GPU} for accelerated processing.

\subsection{Data Splitting}
The $train\_test\_split$ function from $Scikit\_Learn$ was used with the $stratify$ parameter to split the dataset while preserving the class proportions. The dataset consisted of a total of 895 rows with a total of 1626 multi-label compositions across the samples. 
\begin{figure}[htbp]
    \centering
    \includegraphics[width=\columnwidth]{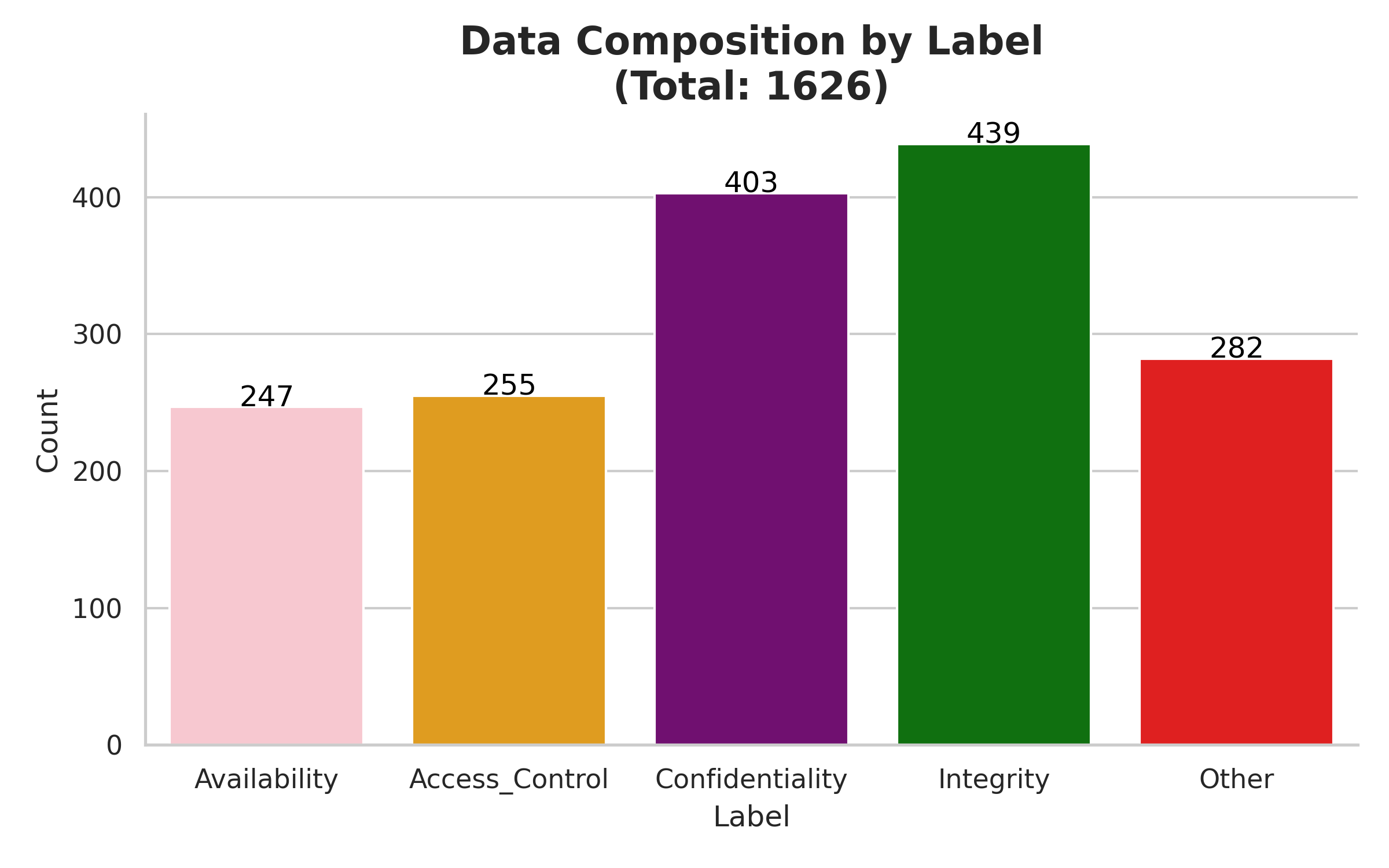} 
    \caption{Data Composition by Label in the CWE Dataset (Total: 1626)}
    \label{fig:data_composition}
        \vspace*{-0.20in}
\end{figure}

The data composition shown in Figure~\ref{fig:data_composition} represents the distribution before stratified sampling. The initial distribution of multi-label classes was done before the dataset was divided. The dataset was first divided into a training set 80\% and a combined validation and test set 20\% while maintaining the proportional distribution of the multi-label classes. The validation and test sets were then further split into 15\% and 5\%, respectively, ensuring that the distribution remained consistent across all subsets. The stratified sampling method made sure that the training, validation, and test sets all had the same proportional class distribution. The dataset comprises five multi-label classes, each representing distinct cybersecurity consequences. The distribution of these labels is as follows: \textit{Availability} (247 instances), \textit{Access Control} (255 instances), \textit{Confidentiality} (403 instances), \textit{Integrity} (439 instances), and \textit{Other} (282 instances). The total number of labeled instances in the dataset is \textbf{1,626}, as illustrated in the figure.

\section{Analysis and Results}
\label{sec:analysis_results}

\subsection{Label-Wise Performance}
To provide a deeper insight into the model's performance, evaluation metrics were calculated for each individual label. Table~\ref{tab:label_wise_metrics} reports these metrics, demonstrating that the model achieved the highest F1-scores for Confidentiality and others.

\begin{table}[!htbp]
\centering
\caption{Label-Wise Performance Metrics of BERT Model}
\label{tab:label_wise_metrics}
\begin{tabular}{|l|c|c|c|c|}
\hline
\textbf{Label}         & \textbf{Accuracy} & \textbf{Precision} & \textbf{Recall} & \textbf{F1-Score} \\ \hline
\textbf{Availability}    & 0.9330            & 0.8627             & 0.8980           & 0.8800            \\ \hline
\textbf{Access Control}  & 0.9385            & 0.8833             & 0.9298           & 0.9060            \\ \hline
\textbf{Confidentiality} & 0.9609            & 0.9538             & 0.9394           & 0.9466            \\ \hline
\textbf{Integrity}       & 0.9050            & 0.8125             & 0.9123           & 0.8595            \\ \hline
\textbf{Other}           & 0.9665            & 0.9625             & 0.9625           & 0.9625            \\ \hline
\end{tabular}
    \vspace*{-0.15in}
\end{table}

From the table, it is evident that the \textbf{Confidentiality} and \textbf{Other} labels exhibit the highest F1-scores, indicating superior model performance in predicting these categories. The \textbf{Access Control} label also performed commendably, with an F1-score of 0.9060. However, the \textbf{Integrity} label showed a relatively lower F1-score, suggesting room for improvement in this area. The observed improvements are primarily attributed to the enhanced contextual understanding of BERT, which is crucial for interpreting cybersecurity texts that contain domain-specific terminology and complex sentence structures.

\begin{figure*}[!htbp]
    \centering
    \begin{subfigure}[b]{0.45\textwidth}
        \centering
        \includegraphics[width=\textwidth, height=5cm]{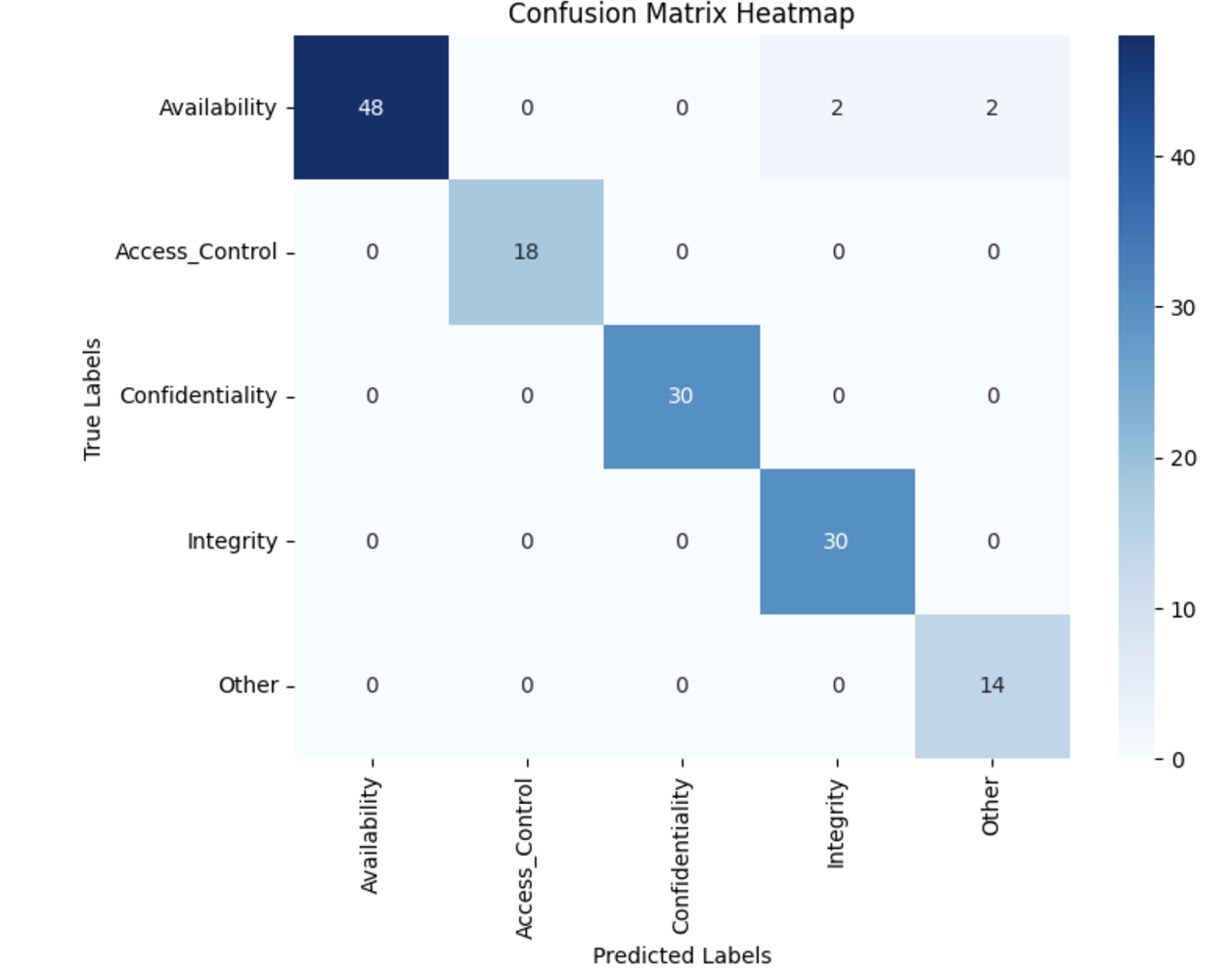}
        \caption{BERT Confusion Matrix}
        \label{fig:bert_conf}
    \end{subfigure}
    \hspace{0.05\textwidth} 
    \begin{subfigure}[b]{0.45\textwidth}
        \centering
        \includegraphics[width=\textwidth, height=5cm]{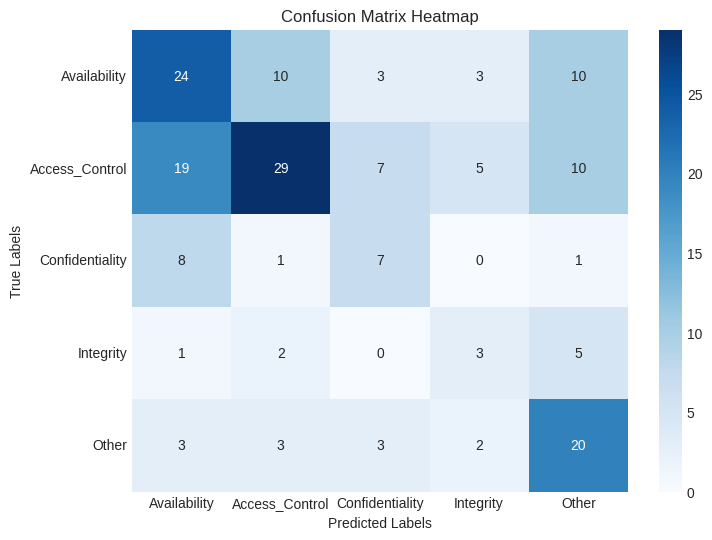}
        \caption{HAN Confusion Matrix}
        \label{fig:han_conf}
    \end{subfigure}
    \caption{Confusion Matrices of Models on Five Labels}
    \label{fig:confusion_matrix}
        \vspace*{-0.20in}
\end{figure*}

\subsection{Confusion Matrix and Comparative Analysis}
As shown in Figure \ref{fig:confusion_matrix}, both the BERT model and the HAN was evaluated using confusion matrices. We analyzed the confusion matrix to gain further insight into the model's classification performance. BERT model \ref{fig:bert_conf} demonstrates exceptional accuracy, correctly classifying 48 instances of availability with only four misclassifications, while access control, confidentiality, integrity, and others were classified with perfect accuracy, showing zero misclassifications. 

In contrast, the HAN model \ref{fig:han_conf} struggles with high misclassification rates, which is evident from the figure. Correctly identified 24 instances of availability but misclassified 26 others, primarily confusing them with Access Control 10 and Other 10. The Access Control category also performed poorly, with 29 correct predictions but 41 misclassifications, often mistaken for Availability 19 times. The figure further highlights HAN’s difficulty in classifying Confidentiality and Integrity, with only 7 and 3 correct classifications, respectively, and frequent misclassifications into other categories. Even the Other category, which had 20 correct classifications, still showed significant confusion across multiple labels.

This suggests that BERT effectively distinguishes between categories, handling the classification task with remarkable precision, as illustrated in the figure. Overall, the figure confirms that BERT significantly outperforms HAN, making far fewer errors and classifying most categories correctly. 


\section{Performance Comparison}
\label{sec:Comparison}

In this section, we compare the performance of our BERT-based model with the results obtained in the original paper\cite{datta2022can}, which used LSTM and CNN architectures for multi-label classification of cyberattack consequences. The comparison focuses on key metrics, including accuracy, precision, recall, and F1-score, for the five predicted labels: Availability, Access Control, Confidentiality, Integrity, and Other.

\subsection{Overall Performance Comparison}
The original paper\cite{datta2022can} explored multiple architectures, including LSTM with multiple outputs, LSTM with a single output, CNN, CNN-LSTM, and CNN-BiLSTM. Table~\ref{tab:original_vs_our_overall} shows the overall performance of the best model in the original paper (CNN-LSTM with random search) compared to our BERT-based model. Our BERT-based model outperformed the best model from the original paper in all metrics, particularly the F1-score.

\begin{table}[!htbp]

    \centering
    \caption{Overall Performance Comparison}
    \label{tab:original_vs_our_overall}
    \scalebox{1.2}{
    \begin{tabular}{|l|c|c|c|c|}
    \hline
    \textbf{Metric} & \textbf{CNN-LSTM (Original\cite{datta2022can})} & \textbf{BERT (Ours)} \\ \hline
    \textbf{Accuracy}   & 0.4357 & 0.9722 \\ \hline
    \textbf{Precision}  & 0.64 & 0.9864 \\ \hline
    \textbf{Recall}     & 0.68 & 0.9731 \\ \hline
    \textbf{F1-Score}   & 0.72 & 0.9797 \\ \hline
    \end{tabular}
    }
        \vspace*{-0.10in}
\end{table}

As shown in Table~\ref{tab:original_vs_our_overall}, our model achieves a significantly higher accuracy of 0.9722 compared to the 0.4357 obtained by the CNN-LSTM architecture. Similarly, our BERT model significantly improves precision, recall, and F1 score, demonstrating its superior ability to capture the relationships between labels.


\subsection{Label-Wise Performance Comparison}
Our models were evaluated on five cybersecurity consequence labels: availability, access control, confidentiality, integrity, and other—using precision, precision, recall, and F1 score as key evaluation metrics. Table \ref{tab:original_vs_our_label_wise} presents a comprehensive comparison of three models: the original CNN-LSTM-RS model and our fine-tuned BERT and HAN models for predicting the consequences of cyberattacks.

\begin{table*}[!htbp]
\centering
\small
\caption{Label-Wise Performance Comparison}
\label{tab:original_vs_our_label_wise}
\resizebox{\textwidth}{!}{ 
\begin{tabular}{|c|c|c|c|c|}
\hline
\textbf{Label}         & \textbf{Metric} & \textbf{Original Paper (CNN-LSTM-RS) \cite{datta2022can}} & \textbf{BERT (Ours)} & \textbf{HAN (Ours)} \\ \hline
\textbf{Availability} & Accuracy  & 0.681  & 0.933  & 0.773  \\ \cline{2-5}
                                      & Precision & 0.53   & 0.86  & 0.61   \\ \cline{2-5}
                                      & Recall    & 0.36   & 0.90  & 0.61   \\ \cline{2-5}
                                      & F1-Score  & 0.42   & 0.88  & 0.61   \\ \hline
\textbf{Access Control} & Accuracy & 0.765  & 0.939  & 0.778  \\ \cline{2-5}
                                      & Precision & 0.56   & 0.88  & 0.64   \\ \cline{2-5}
                                      & Recall    & 0.49   & 0.93  & 0.67   \\ \cline{2-5}
                                      & F1-Score  & 0.52   & 0.91  & 0.65   \\ \hline
\textbf{Confidentiality} & Accuracy & 0.703  & 0.961  & 0.758  \\ \cline{2-5}
                                         & Precision & 0.73   & 0.95  & 0.79   \\ \cline{2-5}
                                         & Recall    & 0.56   & 0.94  & 0.87   \\ \cline{2-5}
                                         & F1-Score  & 0.63   & 0.95  & 0.83   \\ \hline
\textbf{Integrity} & Accuracy  & 0.720  & 0.905  & 0.758  \\ \cline{2-5}
                                      & Precision & 0.78   & 0.81  & 0.77   \\ \cline{2-5}
                                      & Recall    & 0.64   & 0.91  & 0.83   \\ \cline{2-5}
                                      & F1-Score  & 0.70   & 0.86  & 0.80   \\ \hline
\textbf{Other} & Accuracy  & 0.787  & 0.967  & 0.758  \\ \cline{2-5}
                                      & Precision & 0.64   & 0.96  & 0.59   \\ \cline{2-5}
                                      & Recall    & 0.83   & 0.96  & 0.56   \\ \cline{2-5}
                                      & F1-Score  & 0.72   & 0.96  & 0.57   \\ \hline \hline
\textbf{Overall} & Accuracy  & 0.436  & 0.972  & 0.444  \\ \cline{2-5}
                                      & Precision (Micro)  & 0.64  & 0.99  & 0.71  \\ \cline{2-5}
                                      & Recall (Micro)  & 0.68  & 0.97  & 0.75  \\ \cline{2-5}
                                      & F1-Score (Micro)  & 0.72  & 0.98  & 0.73  \\ \cline{2-5}
                                      & F1-Score (Macro)  & -  & 0.98  & 0.69  \\ \hline 
\end{tabular}
}
    \vspace*{-0.20in}
\end{table*}

The BERT-based model demonstrates significant improvements over CNN-LSTM-RS, achieving an overall accuracy of 0.972, which is a substantial increase from the 0.436 reported in the original study. Its strongest performance gains are observed in Confidentiality and Other, where it attains F1-scores of 0.95 and 0.96, respectively. BERT's transformer-based architecture enables it to effectively capture intricate contextual relationships in cybersecurity text, which enhances classification accuracy across all labels. By leveraging pre-trained embeddings and deep contextual representations, BERT effectively models the semantic dependencies between cyber vulnerabilities and their consequences, surpassing the capabilities of traditional deep learning models such as CNN and LSTM.

The HAN model, while not as dominant as BERT in overall classification, exhibits notable strengths in specific labels. With an overall accuracy of 0.444, HAN does not outperform BERT, but it exceeds CNN-LSTM-RS in the Confidentiality and Integrity categories. This suggests that HAN’s hierarchical attention mechanisms are particularly effective for
cybersecurity text analysis, where the data exhibits structural and contextual dependencies. By processing key textual components and emphasizing essential phrases and dependencies, the HAN model effectively captures hierarchical relationships and salient features more effectively than sequence-based models like LSTM.

The confidential category score of 0.95 is a significant leap over the 0.63 achieved by CNN-LSTM-RS, further emphasizing the effectiveness of transformer models in extracting meaningful cybersecurity features from text descriptions. 
These results highlight the superiority of BERT in predictive accuracy when compared to conventional deep learning architectures that rely on sequential modeling.

Despite HAN's moderate overall performance across all five labels, it delivers noteworthy improvement in two key categories: confidentiality and integrity, where its F1-scores surpass those of CNN-LSTM-RS. This suggests that HAN’s hierarchical structure and attention-based framework are particularly well-suited for specific classification tasks, especially when structured text representations are crucial. Although BERT remains the most effective model for classifying general cybersecurity consequences, the targeted strengths of HAN suggest its potential for hybrid approaches, where its hierarchical capabilities could be leveraged in conjunction with state-of-the-art transformer models to enhance cybersecurity text classification.



\section{Conclusion and Future Work}
\label{sec:conclusion}
The paper presents the BERT-based Model and HAN model for predicting cyberattacks with five labels (i.e., Availability, Access Control, Confidentiality, Integrity, and Other) and two labels (i.e., Access Control and Integrity), respectively. In five labels, the BERT model achieved an accuracy of 0.972, precision of 0.99, recall of 0.97, and F1 score of 0.98.  The HAN model achieved an 0.44 in accuracy, 0.71 in precision, 0.75 in recall and 0.73 in F1 score.  The BERT-based model for predicting cyberattack consequences demonstrates significant improvements over previous methods CNN-LSTM-RS \cite{datta2022can}, particularly in terms of precision, recall, F1 score and accuracy. 


The comparison between the original paper and our approach demonstrates the significant improvement brought by the BERT-based model. The original paper employed a combination of CNN and LSTM architectures, which struggled to capture complex relationships in text data. In contrast, the BERT model's pre-trained transformer architecture is better suited for handling long-range dependencies in text, resulting in improved generalization and enhanced multi-label classification performance. Furthermore, using the BERT tokenizer and embeddings also helped people understand the descriptions better by providing more context, which led to higher F1-scores and overall metrics. The superior handling of complex cyberattack descriptions by the BERT model resulted in more accurate predictions of attack consequences across all five labels.\\
\hspace*{1em}Despite the strong performance, the model has certain limitations. One notable challenge is data imbalance, where the \textit{Integrity} label exhibited relatively lower performance, likely due to an uneven distribution of labels in the dataset. Techniques such as data augmentation) to balance the dataset and enhance performance for minority labels. Another limitation is related to long sequence handling, as the BERT model is constrained to processing sequences of up to 256 tokens. This poses a challenge for cybersecurity descriptions that exceed this length. Future research directions include exploring other transformer-based models, such as Alberta, TinyBert, and Roberta, and employing transfer learning techniques to enhance performance on smaller, imbalanced datasets.

\section*{Acknowledgment} 
This work is partially supported by a grant from the National Science Foundation (Award No. 2319802).

\bibliography{refs}{}
\bibliographystyle{plain}

\end{document}